\documentclass[runningheads]{llncs}
\usepackage{graphicx}

\usepackage{amsmath,amsfonts}
\usepackage{algorithmic}
\usepackage{array}
\usepackage{textcomp}
\usepackage{url}
\usepackage{verbatim}
\usepackage{threeparttable}
\begin{document}
\title{An Unsupervised Short- and Long-Term Mask Representation for Multivariate Time Series Anomaly Detection}
%
%
\author{
Qiucheng Miao\inst{1} \and
Chuanfu Xu\inst{1} \and
Jun Zhan\inst{1} \and
Dong Zhu\inst{1} \and
Chengkun Wu\inst{1}
}
\authorrunning{Q. Miao et al.}
%

\institute{$^1$Institute for Quantum Information \& State Key Laboratory of High Performance Computing, College of Computer Science and Technology,
	National University of Defense Technology,
	Changsha, Hunan, P.R. China\\
\email{miaoqiucheng19@nudt.edu.cn}\\}

\maketitle              
\begin{abstract}
Anomaly detection of multivariate time series is meaningful for system behavior monitoring. This paper proposes an anomaly detection method based on unsupervised \textbf{S}hort- and \textbf{L}ong-term \textbf{M}ask \textbf{R}epresentation learning(SLMR). The main idea is to extract short-term local dependency patterns and long-term global trend patterns by using multi-scale residual dilated convolution and Gated Recurrent Unit(GRU) respectively. Furthermore, our approach can comprehend temporal contexts and feature correlations by combining spatial-temporal masked self-supervised representation learning and sequence split. It considers the importance of features is different, and we introduce the attention mechanism to adjust the contribution of each feature. Finally, a forecasting-based model and a reconstruction-based model are integrated to focus on single timestamp prediction and latent representation of time series. Experiments show that the performance of our method outperforms other state-of-the-art models on three real-world datasets. Further analysis shows that our method is good at anomaly localization.

\keywords{Anomaly detection \and Multivariate time series \and Representation learning \and Multi-scale convolution.}
\end{abstract}

\section{Introduction}
Cyber-physical systems are generally used to control and manage industrial processes in some significant infrastructures. Active monitoring of sensor readings and actuator status is crucial for early system behavior detection\cite{10.1145/3447548.3467137}. Anomaly detection is widely studied in different fields to find significant deviations data from normal observations\cite{hundman2018detecting}, such as images and time series. This paper focuses on the anomaly detection of multivariate time series data (MTS). Nevertheless, the data is collected from interconnected sensor networks, so there are usually few labeled anomalies. Therefore, unsupervised learning seems to be the ideal choice for anomaly detection.

Many kinds of research based on machine learning have been proposed to concern MTS anomaly detection\cite{zhang2019deep}\cite{huang2019dsanet}, such as One-class and isolation forest\cite{liu2008isolation}. However, a more generally used strategy is residual-error based anomaly detection. Specifically, the residual-error based anomaly detection relies on a forecasting-based model to predict future sensor measurements\cite{deng2021graph}, or a reconstruction-based model (such as autoencoder) to capture a lower-dimensional representation of sensor measurements[4]. Then the forecasting or reconstruction measurements are compared with the ground-truth measurements to yield a residual error. A system is considered abnormal if the residual error exceeds a threshold.

\begin{figure}[h]
  \centering
  \includegraphics[width=8.5cm]{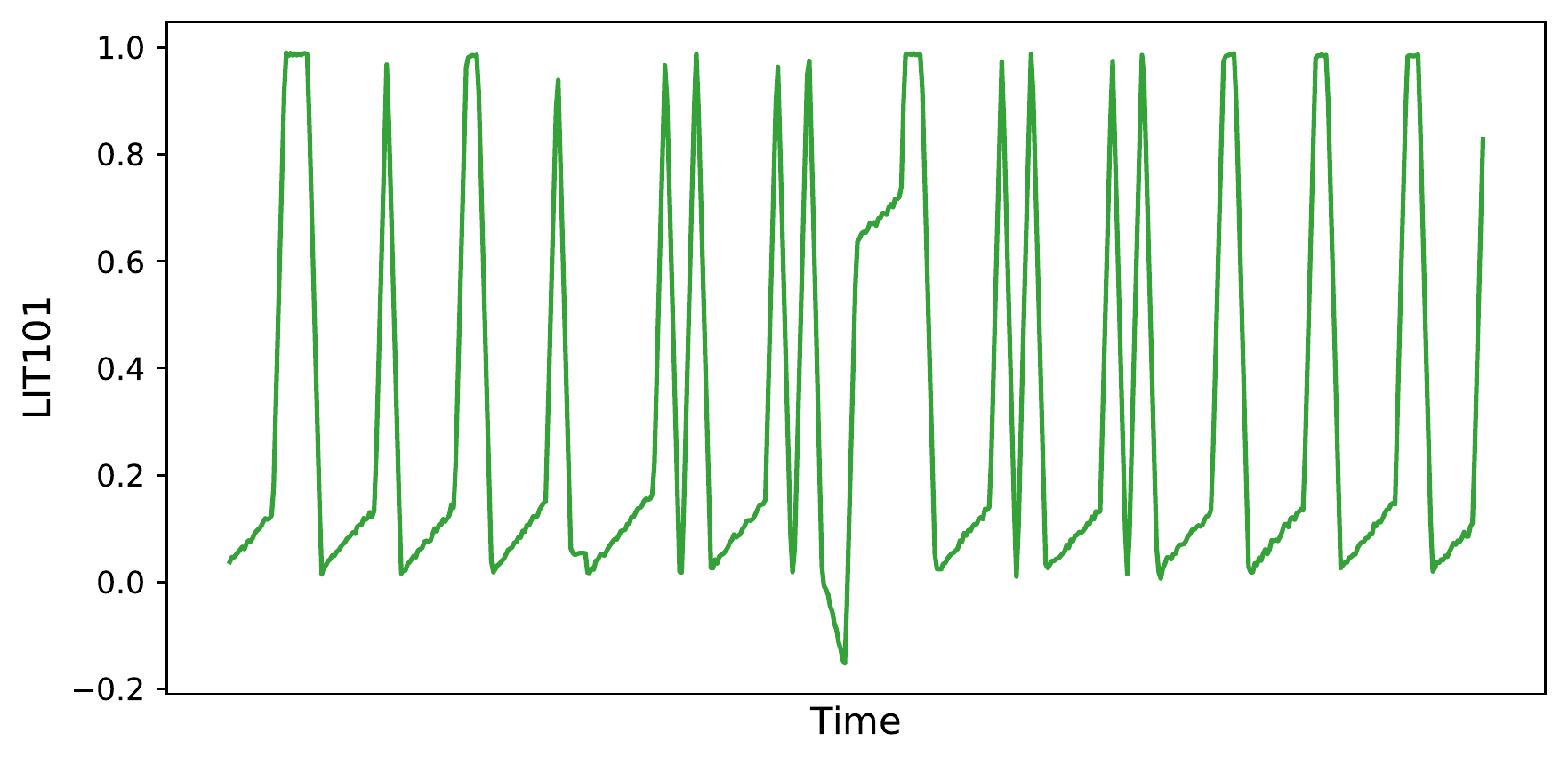}
  \caption{LIT101 state values(The dataset of SWaT)}
\label{tab:fig0}
\end{figure}
\vspace{-2em}

However, the existing methods often fail to consider that time series have different time scales, namely short-term and long-term patterns. Fig.\ref{tab:fig0} shows the state values of LIT101 in the real-world dataset\cite{goh2016dataset}. Obviously, there are two patterns, which are the long-term overall trends and short-term fluctuations. A robust anomaly detection model should be able to capture series correlations on time scales. The long-term patterns reflect the overall trend, and the short-term patterns capture the subtle changes in local regions. Furthermore, MTS is a special kind of sequence data\cite{liu2021time}. Most working condition information can be retained when downsampling the sequence. Sequence split guarantees the model learns an efficient representation with different resolutions. Finally, MTS is composed of a collection of univariate time series, each of which describes an attribute of the system. Therefore, MTS not only has time dependence within the feature, which characterizes the temporal pattern but has inter-feature dependence within the system, which characterizes the linear or non-linear relationship between the features of a system in each period\cite{zhao2020multivariate}. A key concern in anomaly detection is effectively extracting the temporal context and features correlation.

Therefore, we propose a jointly optimized anomaly detection method based on short- and long-term mask representation. The main contributions are the following:

(1) Random mask and sequence split: The input data is masked randomly and then the original data is used for reconstruction and forecasting representation. Mask can promote the model to understand temporal contexts and learn the dynamic information between features. In addition, the input data is split to obtain odd subsequences and even subsequences. Different convolution filters are used to extract the features, maintain heterogeneity information and ensure that the model learns different sequence resolutions.

(2) Short- and long-term patterns extract: We perform multi-scale residual dilated convolution to extract short-term spatial-temporal local dependency patterns and filter irrelevant information. Also, an attention mechanism is applied to different channels, adjusting the contribution of different feature weights. Finally, the jointly optimized method based on GRU is introduced to identify long-term patterns for time series trends. The forecasting-based model focuses on single timestamp prediction, while the reconstruction-based model learns the latent representation of the entire time series.

The results illustrate that our method is generally better than other advanced methods, and more importantly, achieves the best F1 score on the three datasets. It also shows that the proposed method enables locating anomalies.


\vspace{-1em}
\section{Related work}

There is plenty of literature on time-series anomaly detection, which can be classified into two categories. One is the forecasting-based model and another is the reconstruction-based model. 

The forecasting-based model implements the prediction of the next timestamp, which is compared with the ground-truth value to generate a residual error, to decide whether an abnormality occurs according to the threshold. Long Short-Term Memory (LSTM) is one of the most popular methods to predict the next time sequence in anomaly detection\cite{hundman2018detecting}. The attention mechanism and its variants have also obtained widespread application, such as Dsanet\cite{huang2019dsanet}. Building a model, the complex dependencies of features, is a great challenge in traditional machine learning, so GNN-based anomaly detection has become a hot research topic\cite{deng2021graph}\cite{zhao2020multivariate}. CNN can automatically extract and learn the different components of the signal on time scales, which yield unusually brilliant results in multivariate time series anomaly detection\cite{ren2019time}. 

The reconstruction-based model learns the latent representation of the entire time series by reconstructing the original data to generate the observed value. Then, the deviation between the observed value and the ground-truth value is evaluated to identify anomalies. The current state-of-the-art deep reconstruction anomaly detection models mainly include DAGMM\cite{zong2018deep}, AE\cite{su2019robust}, GAN\cite{li2019mad}. In addition, convolutional neural networks perform well in feature extraction, especially in noisy environments\cite{zhang2019deep}\cite{9507359}.


\section{Methodology}
\textbf{Problem definition:} The dataset of MTS anomaly detection can be expressed as $\boldsymbol{S}\in \mathbb{R}^{n\times k}$, where $n$ is the length of the timestamp, and $k$ is the number of input features. A fixed input $\boldsymbol{X}\in \mathbb{R}^{w\times k}$ is generated from long time series by a sliding window of length $w$. The target of our algorithm is to produce a set of binary labels $\boldsymbol{y} \in \mathbb{R}^{n-w}$, where $y(t)=1$ indicates the $t_{th}$ timestamp is abnormal.


\begin{figure}[h]
  \centering
  \includegraphics[width=9cm]{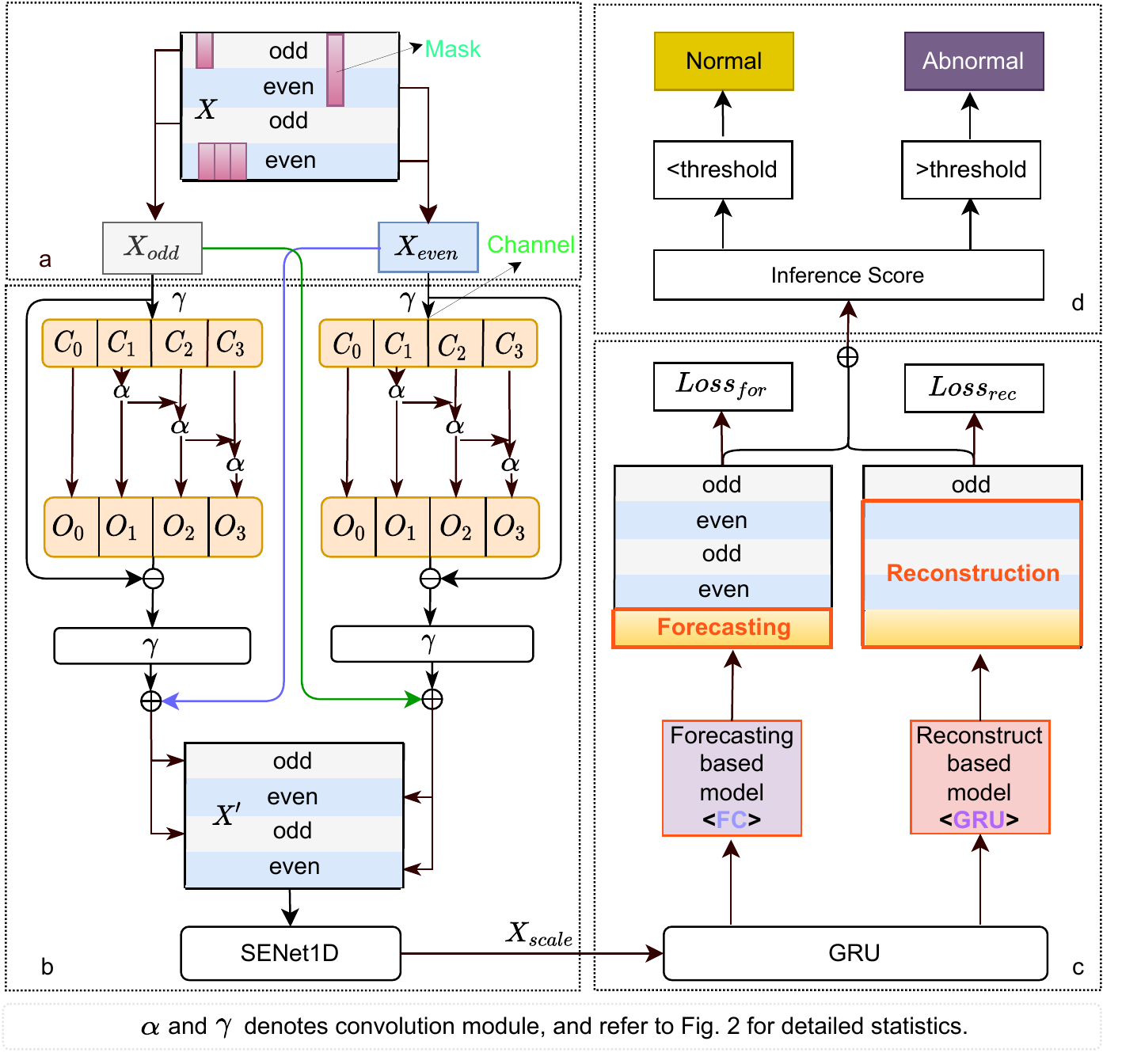}
  \caption{An overview of our method for anomaly detection. a) We randomly set part of the input data to 0, then divide the input into odd and even subsequences according to the time dimension. b) After odd and even subsequences are generated, firstly, we realize the channel dimension transformation with a $1 \times1$ convolution (i.e., $\gamma$, see Fig. \ref{tab:fig2}), and then perform multi-scale residual convolution on different channels (i.e., $\alpha$, see Fig. \ref{tab:fig2}). The output channel is inverse transformation through a $1 \times1$ convolution, merging odd and even sequences. Then all of them are input into SENet1D (see Fig. \ref{tab:fig3}), and the contribution weights of different channels are adjusted. c) GRU-based sequence joint optimization, and balance the scores of both. d) Determine whether an abnormality occurs according to the threshold and the final inference score.}
\label{tab:fig1}
\end{figure}
\vspace{-2em}

\subsection{Mask and sequence split}
As a task for spatial-temporal masked self-supervised representation, the mask prediction explores the data structure to understand the temporal context and features correlation. We will randomly mask part of the original sequence before we input it into the model, specifically, we will set part of the input to 0. The training goal is to learn the latent representation of the masked signal and then predict the masked value. The mask prediction task involves the basic assumption that there is a correlation between temporal context and features. The benefits are twofold: First, masked values do not change the size of the time series, which is essential for time series anomaly detection. Second, they also improve the robustness of the learned representation by forcing each timestamp to reconstruct itself in a different context. Intuitively, masked values enable the network to infer the representation in an incomplete time series, which helps predict missing values in incomplete surrounding information.

Expressly, part of the input is set to 0, and the model is required to predict the masked values, as shown in Fig. \ref{tab:fig1}(a). A binary mask $\boldsymbol{M}\in \mathbb{R}^{w\times k}$ for each independent sample is firstly created, and then do the element-wise product with the input $\boldsymbol{X}: \boldsymbol{\widetilde{X}}=\boldsymbol{M}\odot \boldsymbol{X}$. $\boldsymbol{M}$ is referenced by the following method, a transformer-based framework\cite{10.1145/3447548.3467401}. For each $\boldsymbol{M}$, a binary mask is alternately set, and the length of each mask segment (0 sequences) obeys the geometric distribution probability $1/s_m$, which is the probability that each masked sequence will stop. And unmask segment (1 sequence) obeys the geometric distribution probability $\frac{1}{s_m}\times\frac{r}{(1-r)}$, which is a probability that each unmasked sequence will stop. $r$ is the ratio of alternating 0 and 1 in each sequence. We chose the geometric distribution because for time series, the powerful latent variable representation capability of the deep model allows adaptive estimation of short masked sequences (which can be considered as outliers). Therefore, we need to set a higher proportion of masked sequences to force the model actively mine contextual associations. The value of $r$ has an impact on the results refer to Fig. \ref{tab:fig4}. We set $s_m=3$ and $r=0.1$, which has the best performance with extensive experiments.

The original time-series is split, and the input sequence $\boldsymbol{X}\in \mathbb{R}^{w\times k}$ is divided into even sequence $\boldsymbol{X}_{even}\in \mathbb{R}^{\left(w/2\right)\times k}$ and odd sequence $\boldsymbol{X}_{odd }\in \mathbb{R}^{\left(w/2\right)\times k}$, each of which temporal resolution is coarser than the original input. The subsequence only contains part of the original information, which well preserves the heterogeneous information of the original sequence. Feature extraction is performed on the odd-sequence and even-sequence, respectively. Interactive learning between features is added to each sequence to compensate for the representation loss during downsampling. As shown in Fig. \ref{tab:fig1}(a), after projecting $\boldsymbol{X}_{even}$ and $\boldsymbol{X}_{odd}$ to two different one-dimensional convolutions $\gamma$ and $\alpha$, the $\boldsymbol{X}_{even}^\prime$ and $\boldsymbol{X}_{odd}^\prime$ are obtained by the residuals connection. After all downsampling---convolution---interaction, all low-resolution components are rearranged and connected to a new sequence representation.

\begin{equation}
  \boldsymbol{X}_{even}^\prime=\boldsymbol{X}_{odd}\oplus Conv1d(\boldsymbol{X}_{even})
\end{equation}
\begin{equation}
  \boldsymbol{X}_{odd}^\prime=\boldsymbol{X}_{even}\oplus Conv1d(\boldsymbol{X}_{odd})
\end{equation}

Conv1d denotes the multi-scale residual convolution operation, and $\oplus$ is the residual connection. Time series downsampling can retain most information and exchange information with different time resolutions. In addition, the designed sequence sampling does not require domain knowledge and can be easily generalized to various time-series data.

\subsection{Multi-scale	residual convolution}
\begin{figure}[h]
  \centering
  \includegraphics[width=8cm]{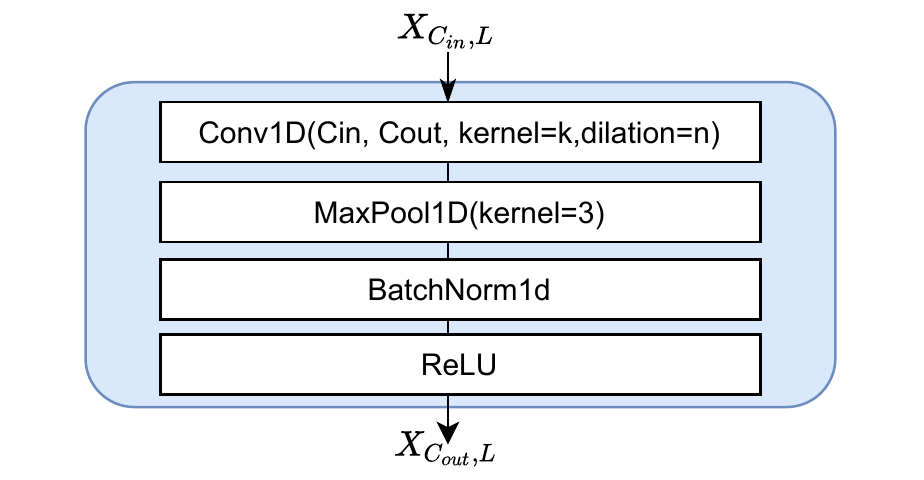}
  \caption{Convolution modules, where $\alpha$: k=2i+1, (i=1,...,s-1), dilation=2, $\gamma$: k=1, dilation=1, $\beta$: k=3, dilation=1.}
	\label{tab:fig2}
\end{figure}

CNN uses the learnable convolution kernels to automatically extract features from different scales to obtain a better representation\cite{gao2019res2net}. Therefore, we propose a simple and effective module for multi-scale residual convolution. Unlike existing multi-layer and multi-scale methods, we have improved the multi-scale representation ability of CNN at a more fine-grained level. To achieve this goal, firstly, realizing the feature channels dimension($n$) transformation through $1\times1$ convolution ($\gamma$, as shown in Fig.\ref{tab:fig2}). Then, the feature channels will be divided into $s=4$ subsets on average, each subset can be represented by $\boldsymbol{C}_i,\left(i=1,2,..,s\right)$, and each subset of the segmented has $w$ channels, i.e., $\boldsymbol{C}_i \in \mathbb{R}^{\frac{n} {2} \times w}$, where $w$ denotes the number of feature map subgroups. A group of $3\times3$ kernels extracts features from a set of input feature maps, denoted by $\alpha$, as shown in Fig.\ref{tab:fig2}. Finally, the output features of the previous group are connected with another group in a residual manner, and the kernel size of this group is $2i+1, i=0,1,...,s-1$. This process is repeated several times until the feature maps of all groups are connected and concatenate all outputs together.

The output $\boldsymbol{O}_i$ can be expressed as:

\begin{equation}
\boldsymbol{O}_{i}=\left\{\begin{array}{c}
\boldsymbol{C}_{i},i=1 \\
\alpha\left(\boldsymbol{C}_{i}\right),i=2 \\
\alpha\left(\boldsymbol{C}_{i}+\boldsymbol{O}_{i-1}\right), 2<i \leq s
\end{array}\right.
\end{equation}

$\alpha$ is a convolution module, as shown in Fig\ref{tab:fig2}. The multi-scale residual can capture a larger receptive field. To effectively extract local and global information, different convolution scales are integrated. The $1\times1$ kernel convolution ($\gamma$) adjusts the output data to the same size as the input, as shown in Fig.\ref{tab:fig2}.

\begin{figure}
  
  \centering
  \includegraphics[width=8cm]{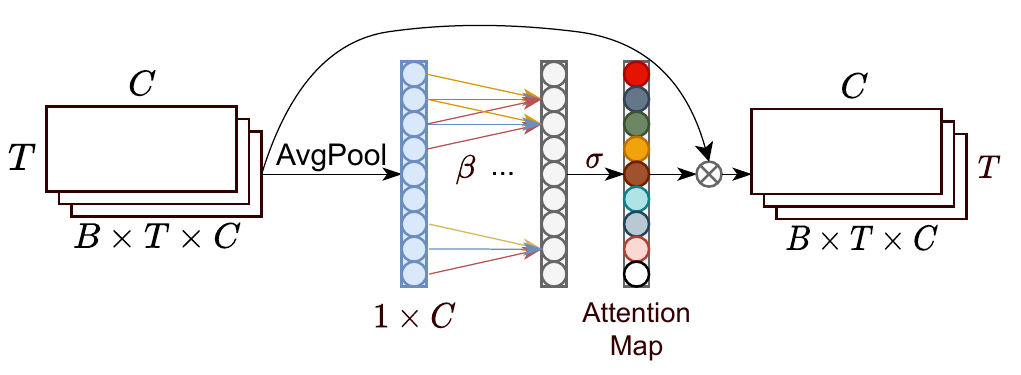}
  \caption{SENet1D module. $\beta$ denotes convolution with a kernel size of 3, $\sigma$ is the Sigmoid activation function, $\otimes$ is element-wise broadcast dot multiplication.}
	\label{tab:fig3}
\end{figure}

In different feature channels, the contribution in the anomaly detection is often different, and SENet1D is used to adjust the weight contribution of different features, as shown in Fig. \ref{tab:fig3}.  The specific path to obtain weights is "Global Average Pooling (AvgPool) $\to$ Conv1d $\to$ ReLU Function $\to$ Conv1d $\to$ Sigmoid Function ($\sigma$)", as shown in Fig. \ref{tab:fig3}.

\begin{equation}
\boldsymbol{z}=\sigma(Conv1d(\delta(Conv1d(Avgpool(\boldsymbol{X}^\prime)))))
\end{equation}
\begin{equation}
\boldsymbol{X}_{scale}=\boldsymbol{z}\otimes \boldsymbol{X}^\prime
\end{equation}

Where $\boldsymbol{X}^\prime$ is the original sequence sample, $\delta$ represents the ReLU function, Conv1d is a one-dimensional convolution with a kernel size of 3, which implements a nonlinear gating mechanism. Avgpool is a global average pooling, which compresses the global spatial information to the channel descriptors to generate channel-oriented statistical information. Finally, through $\sigma$, the attention weights of different channels are compressed to [0,1], realizing the channel weight contribution adjustment, $\boldsymbol{X}_{scale}$ inputs into the GRU to capture the long-term sequence trends.

\subsection{Joint reconstruction and forecasting based on GRU}
The forecasting-based and reconstruction-based models have advantages and complement each other, as shown in Fig. \ref{tab:fig1}(c). 

(1) Forecasting-based model: The forecasting-based model is to predict the next timestamp. We achieve the prediction by fully connected layers after the GRU, and the loss function is the root mean square error (RMSE):

\begin{equation}
{Loss}_f=\sqrt{\sum_{m=1}^{k}{({\hat{x}}_{t+1,m}-x_{t+1,m})}^2}
\end{equation}

${Loss}_f$ represents the loss function of the forecasting-based model. Where ${\boldsymbol{\hat{x}}}_{t+1,m}$ denotes the predicted value of node $m$ at $t+1$, $\boldsymbol{x}_{t+1,m}$ represents the groundtruth value of node $k$ at time $t+1$. By calculating the residual between the predicted value and the ground truth value, it is determined whether this point is abnormal.

(2) Reconstruction-based model: Reconstruction-based model learns low-dimensional representations and reconstructs the “normal patterns” of data. For the input $\boldsymbol{X}_{t-w:t}$, GRU decodes the input to obtain the reconstruction matrix ${\boldsymbol{\hat{X}}}_{t-w:t}$, $w$ is window sizes, and the RMSE is:

\begin{equation}
{Loss}_r=\sqrt{\sum{(\boldsymbol{X}_{t-w:t}-{\boldsymbol{\hat{X}}}_{t-w:t})}^2}
\end{equation}

${Loss}_r$ represents the loss function of the reconstruction-based model. The total loss is the sum of ${Loss}_f$ and ${Loss}_r$:

\begin{equation}
Loss_{total}= Loss_f + Loss_r
\end{equation}

\subsection{Anomaly detection}
For the joint optimization, we have two inference results for each timestamp. One is the predicted value ${{\hat{x}}_i|i=1,2,\ldots,k}$ computed forecasting-based model, and the other is the reconstructed value ${{\hat{x }}_j|j=1,2,\ldots,k}$, which was obtained from reconstruction-based model. The final inference score balances their benefits to maximize the overall effectiveness of anomaly detection, as shown in Fig. \ref{tab:fig1}(d). We use the mean of all features to compute the final inference score and to determine whether the current timestamp is abnormal with the corresponding inference score.

For the forecasting-based model, the input data is $\boldsymbol{X}_{t-w:t}$, and inference score is $\sqrt{\sum_{m=1}^{k}{({\hat{x}}_{t+1,k}-x_{t+1,k})}^2}$. For the reconstruction-based model, the input data is $\boldsymbol{X}_{t-w+1:t+1}$, using GRU decodes the input to obtain the reconstruction matrix ${\boldsymbol{\hat{X}}}_{t-w+1:t+1}$, and inference score is $\sqrt{\sum{({\boldsymbol{\hat{X}}}_{t:t+1}-\boldsymbol{X}_{t:t+1})}^2}$. The final inference score can be calculated by:
\begin{equation}
score =  \frac{1}{k} (\sqrt{\sum_{m=1}^{k}{({\hat{x}}_{t+1,m}-x_{t+1,m})}^2} + \gamma \sqrt{\sum{({\boldsymbol{\hat{X}}}_{t:t+1}-\boldsymbol{X}_{t:t+1})}^2})
\end{equation}

The optimal global threshold selection is similar to previous work, we also use best-F1\cite{su2019robust}\cite{audibert2020usad}\cite{goh2017anomaly}\cite{zhao2020multivariate}. It is considered an anomaly when the inference score exceeds the threshold.

\section{Experiment and Analysis}

%
%
%

\subsection{Datasets and evaluation metrics}

\textbf{a) Datasets:} 
To verify the effectiveness of our model, we conduct experiments on three datasets, SMAP (Soil Moisture Active Passive satellite)\cite{hundman2018detecting}, MSL (Mars Science Laboratory rover)\cite{hundman2018detecting}, and SWaT (The Secure Water Treatment)\cite{goh2016dataset}. Table~\ref{tab:table1} is the detail statistic information of the three datasets. Our code with Pytorch1.8 and data are released at \url{https://github.com/qiumiao30/SLMR}.


\begin{table}[]
\centering
\caption{Statistical details of datasets. (\%) is the percentage of abnormal data.}
\label{tab:table1}
\begin{threeparttable}
\begin{tabular}{ccccc}
\hline
\textbf{Dataset} & \textbf{Train}  & \textbf{Test}   & \textbf{Dimensions} & \textbf{Anomalies} (\%) \\ \hline
SWaT    & 480599\tnote{1} & 449919 & 51         & 11.98        \\
SMAP    & 135183 & 427617 & 55*25\tnote{2}      & 13.13          \\
MSL     & 58317  & 73729  & 27*55      & 10.72          \\ \hline
\end{tabular}
\begin{tablenotes}
       \footnotesize
       \item[1] Remove the first four hours of data\cite{li2019mad}.
       \item[2] 55 is the dimension, 25 is the number of entities
     \end{tablenotes}
\end{threeparttable}
\end{table}
\textbf{b) Metrics:} We use precision, recall, and F1 score to measure the performance of our model. In a continuous anomaly segment, the entire segment is correctly predicted if at least one moment is detected to be anomalous. The point-adjust method is applied to evaluate the performance of models according to the evaluation mechanism in \cite{su2019robust}\cite{zhao2020multivariate}\cite{audibert2020usad}. The model is trained with the Adam optimizer, the learning rate is initialized to 0.001, the batch size is 256, 10\% of the training data is used as the validation set, and the window size is set to 100 or 80.

\begin{table*}[]
\centering
\caption{Performance of our model and baselines.}
\label{tab:table2}
\begin{tabular}{cccccccccc}
\hline
{\textbf{Method}} & \multicolumn{3}{c}{\textbf{SWaT}}             & \multicolumn{3}{c}{\textbf{SMAP}}             & \multicolumn{3}{c}{\textbf{MSL}}              \\ \cline{2-10} 
                                 & \textbf{Prec.} & \textbf{Rec.} & \textbf{F1}    & \textbf{Prec.} & \textbf{Rec.} & \textbf{F1}    & \textbf{Prec.} & \textbf{Rec.} & \textbf{F1}    \\ \hline
IF                               & 0.962         & 0.731        & 0.831          & 0.442         & 0.510        & 0.467          & 0.568         & 0.674        & 0.598          \\
AE                               & 0.991         & 0.704        & 0.823          & 0.721         & 0.979        & 0.777          & 0.853         & 0.974        & 0.879          \\
LSTM-VAE                         & 0.979         & 0.766        & 0.860          & 0.855         & 0.636        & 0.729          & 0.525         & 0.954        & 0.670          \\
MAD-GAN                          & 0.942         & 0.746        & 0.833          & 0.671         & 0.870        & 0.757          & 0.710         & 0.870        & 0.782          \\
DAGMM                            & 0.829         & 0.767        & 0.797          & 0.633         & 0.998        & 0.775          & 0.756         & 0.980        & 0.853          \\
LSTM-NDT                         & 0.990         & 0.707        & 0.825          & 0.896         & 0.884        & 0.890          & 0.593         & 0.537        & 0.560          \\
OmniAnomaly                      & 0.722         & 0.983        & 0.832          & 0.758         & 0.975        & 0.853          & 0.914         & 0.889        & 0.901          \\
USAD                             & 0.987         & 0.740        & 0.846          & 0.769         & 0.983        & 0.863          & 0.881         & 0.978        & 0.927          \\
MTAD-GAT                         & 0.903         & 0.821        & 0.860          & 0.809         & 0.912        & 0.901          & 0.875         & 0.944        & 0.908          \\
GTA                         & 0.948         & 0.881        & 0.910          & 0.891         & 0.917        & 0.904          & 0.910         & 0.911        & 0.911          \\
\textbf{SLMR}                & 0.963       & 0.874        & \textbf{0.916} & 0.915         & 0.992        & \textbf{0.952} & 0.965         & 0.967        & \textbf{0.966} \\ \hline
\end{tabular}
\end{table*}
\vspace{-2em}

\subsection{Performance and analysis}
We compare our method with other 10 advanced models that deal with multivariate time series anomaly detection, including Isolation Forest(IF)\cite{liu2008isolation}, DAGMM\cite{zong2018deep}, basic Autoencoder, LSTM-VAE\cite{park2018multimodal}, MAD-GAN\cite{li2019mad}, LSTM-NDT\cite{goh2017anomaly}, USAD\cite{audibert2020usad},  OmniAnomaly\cite{su2019robust},  MTAD-GAT\cite{zhao2020multivariate},  GTA\cite{9497343}. The results illustrate that our method generally achieves the highest F1 score on the three datasets. We can also observe that our method achieves a great balance between precision and recall.

\begin{figure}
  \centering
  \includegraphics[width=10.5cm]{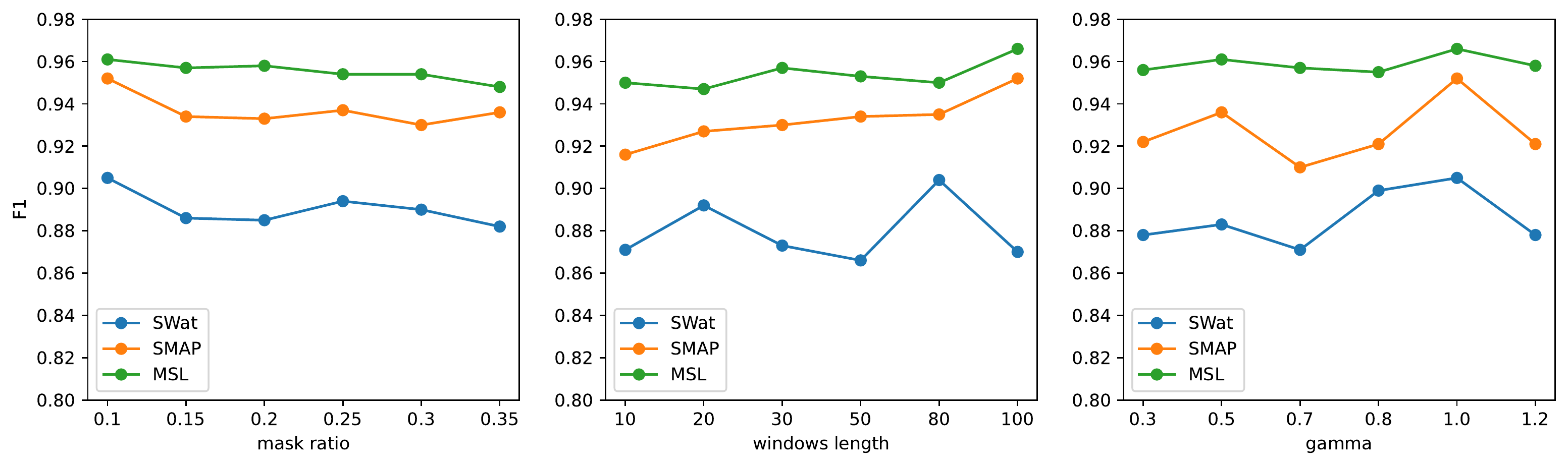}
  \caption{Parameter sensitivity.}
	\label{tab:fig4}	
\end{figure}


\textbf{a). Overview Performance:} Overall, the non-parametric methods, Isolation Forest and DAGMM, perform the worst. Because they are poor at capturing temporal information. As a generative model, DAGMM has high recall scores on the SMAP and MSL datasets, but it still does not solve the time correlation problem, resulting in poor performance. For time series, temporal information is necessary because observations are dependent and historical data helps reconstruct/predict current observations. AutoEncoder(AE) uses an encoder and decoder to complete the reconstruction of time series. USAD adds adversarial training based on AE to generate better representations for downstream tasks. However, the above two methods reconstruct the time series point by point without capturing the time correlation, limiting the model's detection performance. Generative models based on VAE or GAN, such as LSTM-VAE, MAD-GAN, and OmniAnomaly, can effectively capture temporal information but do not consider feature-level correlation. MTAD-GAT and GTA capture feature dependence through graph attention networks, but to a certain extent, the temporal information is ignored. The method proposed in this paper firstly combines mask spatial-temporal representation, which facilitates the model to comprehend temporal contexts and feature correlations. In addition, multi-scale convolution extracts short-term dependency patterns, catches rich information on time scales, filters the original data, and reduces the impact of irrelevant information on the results. Finally, GRU acquires long-term dependencies. Therefore, our method achieves the highest F1 score and the best performance in balancing recall and precision.

\begin{figure}[h]
  \centering
  \includegraphics[width=8.5cm]{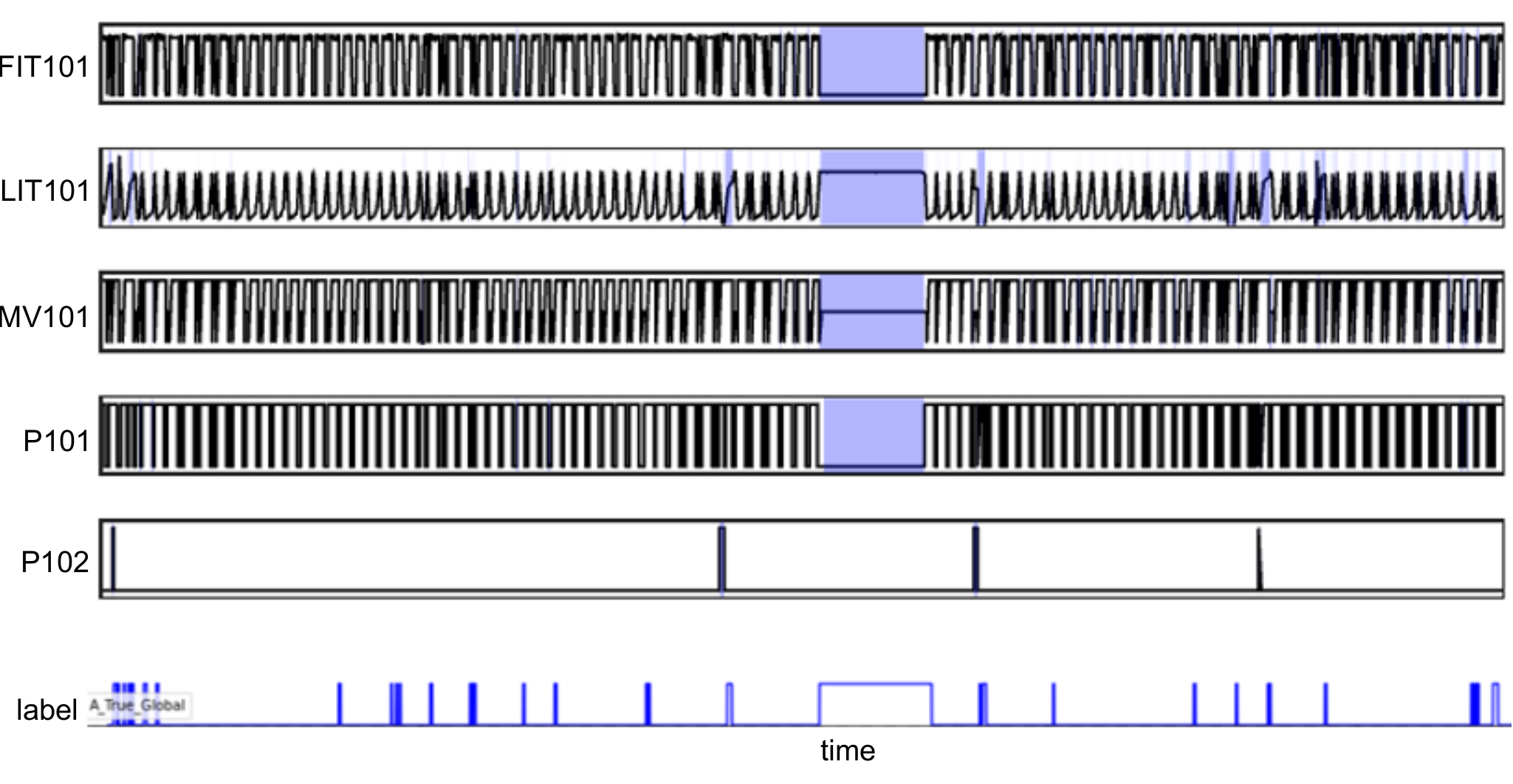}
  \caption{Anomalies localization(the first 5 features). The shadow is the abnormal segment detected by the model, and the 'label' is the true global abnormal segment.}
	\label{tab:fig6}
\end{figure}

\textbf{b). Parameter sensitivity: } In Fig. \ref{tab:fig4}, we demonstrate the sensitivity of key parameters to results. The model can achieve high performance with a mask ratio of 0.1. If the mask ratio is too high, it may cause the rate of missing data to be too high, and thus the model is unable to fit the actual distribution, especially when the dataset size is small. SWaT shows slight fluctuations with the change of windows length. It is possible that the data have obvious periodic characteristics, as shown in Fig. \ref{tab:fig6}, which is more sensitive to windows length. When the gamma is 1, the performance is almost the best, the contribution of the forecasting-based model and the reconstruction-based model are almost similar, and the final inference score is balanced.

\subsection{Ablation study}
We perform ablation studies using several variants of SLMR to further point out the validity of the module described in Section 3, as shown in Table~\ref{tab:table3}.

\textbf{Self-supervised mask representation.} The SWaT and SMAP have been dramatically improved, but the MSL has poor performance. It is proved that when datasets size is small, the performance is not improved. The main reason is that the context learning ability of the model is poor on small-scale datasets, and the masked value cannot be effectively filled.
\textbf{Sequence split.} The data downsampling is beneficial to improving performance. This method can ensure that the model learns the effective representation of the sequence between different levels.
\textbf{Multi-scale residual dilation convolution.} The proposed method can learn more latent knowledge than a basic convolution layers, and it can effectively reduce the impact of irrelevant information on the results.
\textbf{Joint optimization.} The forecasting-based model is sensitive to the randomness of the time series, while the reconstruction-based model alleviates it by learning the distribution of random variables. Besides, the reconstruction-based model can capture the global data distribution well, but it may ignore abrupt perturbations, thereby destroying the periodicity of the time series. In contrast, the forecasting-based model can effectively compensate for this drawback.

\vspace{-1em}
\begin{center}
\begin{table}[]
\centering
\caption{Performance of our method and its variants(F1).}
\label{tab:table3}
\begin{tabular}{lccc}
\hline
\quad{}\textbf{Method}   & \textbf{SWaT}  & \textbf{SMAP}  & \textbf{MSL}   \\ \hline
\quad{} \textbf{SLMR} & \textbf{0.916} & \textbf{0.952} & \textbf{0.966} \\ \hline
\quad{}w/o mask             & 0.854          & 0.942          & \textbf{0.966} \\
\quad{}w/o odd/even         & 0.873          & 0.911          & 0.953          \\
\quad{}w/o multi\_CNN       & 0.849          & 0.929          & 0.951          \\
\quad{}w/o SENet1D          & 0.870          & 0.903          & 0.950          \\
\quad{}w/o forecast         & 0.852          & 0.923          & 0.952          \\
\quad{}w/o reconstruct      & 0.847          & 0.935          & 0.941          \\ \hline
\end{tabular}
\end{table}
\end{center}
\vspace{-2em}

\begin{figure}
  \centering
  \includegraphics[width=9.5cm]{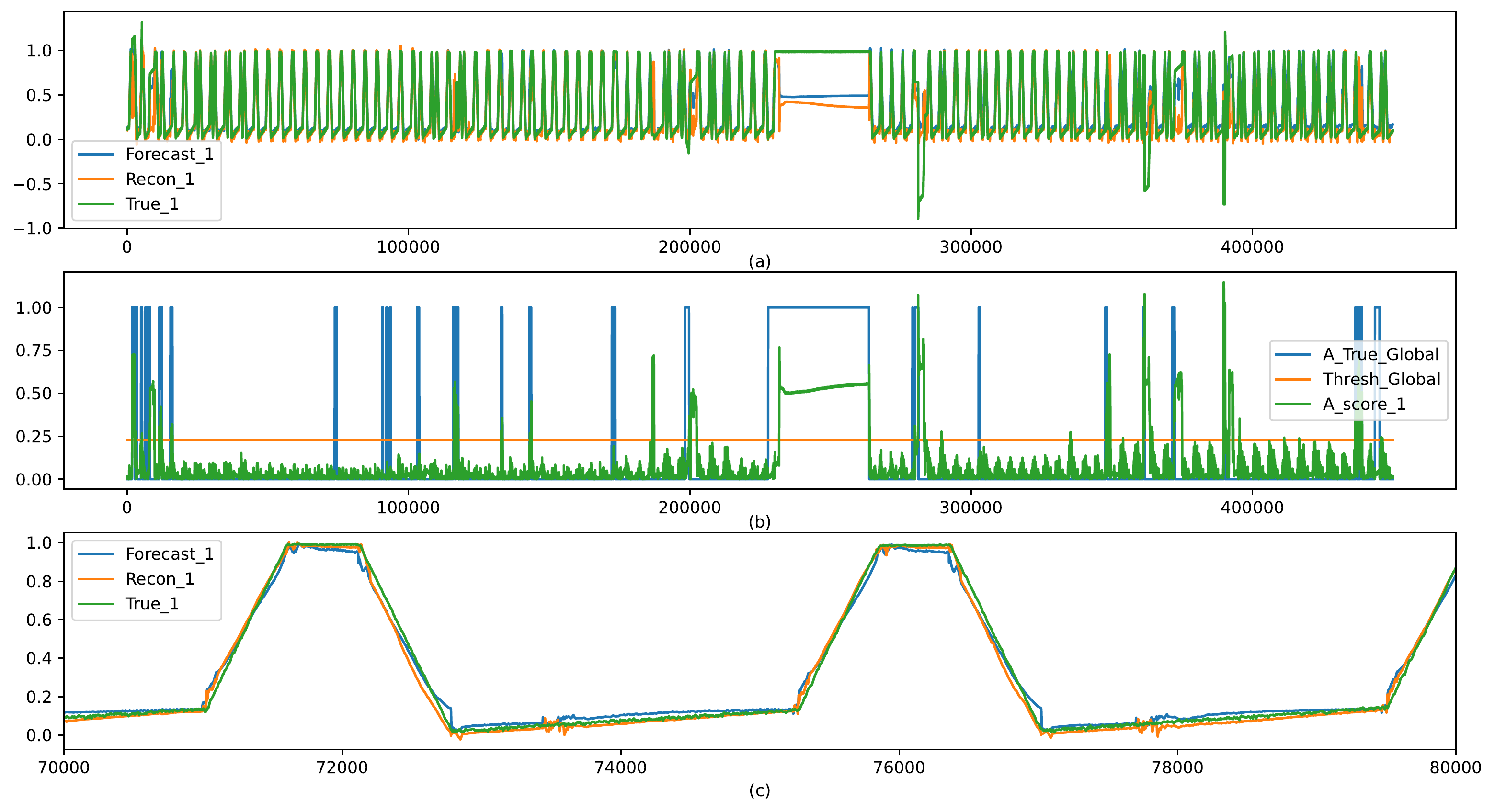}
  \caption{Visualization of LIT101. (a): True\_1 is ground-truth values, and Forecast\_1 and Recon\_1 is generated value by forecasting and reconstruction; (b): A\_True\_Global is the true global anomaly state, and A\_score\_1 is inferred anomaly scores, and Thres\_Global is the global threshold; (c): Partial subset of (a).}
	\label{tab:fig7}
\end{figure}

\subsection{Localization abnormal}
Fig. \ref{tab:fig6} shows the abnormal location information of the SWaT. The blue part is the detected abnormal event. It can be seen that the intermediate abnormal information can be wholly detected. To other shorter abnormal segments, most of the abnormal information is detected. As shown in Fig. \ref{tab:fig7}, the abnormal detection of LIT101 (the green line of (b)) corresponds to the actual measured abnormal state, which fully proves that our model can effectively detect abnormal information. Therefore, accurate anomaly location interpretation is convenient for practitioners to find the abnormal parts early, and timely mitigation measures can be taken to avoid equipment downtime or damage due to major failures and reduce potential economic and environmental losses.

'A\_score\_1' in Fig.\ref{tab:fig7}(b) represents the actual anomaly score inferred from LIT101, which fully proves that our model can effectively detect abnormal information. Therefore, accurate anomaly location interpretation is convenient for the engineer to find the abnormal parts early. They can timely take mitigation measures to avoid equipment outages or damage due to major failures and reduce potential economic and environmental losses.

\section{Conclusion}
This paper proposes anomaly detection methods based on MTS. Firstly, we use mask-based self-supervised representation learning to enable the model to understand temporal contexts and features correlation, and sequence split enhances the representation capacity of different resolutions of sequences. Then, multi-scale residual convolution effectively improves the capability to capture short-term dynamic changes, and filters existing irrelevant information. GRU identifies the long-term trend patterns by joint optimization of the forecasting and reconstruction-based model. It is observed that our method effectively captures the temporal-spatial correlation, and generally outperforms the other advanced methods.

%
%
%
\bibliographystyle{splncs04}
\bibliography{sample-base}
%




\end{document}